\pdfoutput=1
\documentclass[11pt]{article}

\usepackage[preprint]{acl}
\usepackage{times}
\usepackage{latexsym}
\usepackage[T1]{fontenc}
\usepackage[utf8]{inputenc}
\usepackage{microtype}
\usepackage{inconsolata}
\usepackage{graphicx}
\usepackage{booktabs}
\usepackage{multirow}
\usepackage{amsmath}
\usepackage{amssymb}

\usepackage{amsthm}
\newtheoremstyle{named}{\topsep}{\topsep}{\itshape}{0pt}{\bfseries}{.}{.5em}%
  {\thmname{#1}\thmnumber{ #2}\thmnote{. #3}}
\theoremstyle{named}
\newtheorem{proposition}{Proposition}

\usepackage{etoolbox}
\newcommand{\tabunderline}[1]{\setbox0\hbox{#1}\rlap{\rule[-0.9pt]{\wd0}{0.35pt}}\box0}
\AtBeginEnvironment{table}{\let\underline\tabunderline}
\AtBeginEnvironment{table*}{\let\underline\tabunderline}

\usepackage{algorithm}
\usepackage{algpseudocode}

\graphicspath{{figures/}}
\makeatletter
\def\input@path{{figures/}}
\makeatother

\title{PACT: Certifying What Is Written in Parallel Decoding\\
  of Diffusion Language Models}

\author{
 \textbf{Yanchao Li\textsuperscript{1,2}},
 \textbf{Yusu Zhang\textsuperscript{2}},
 \textbf{Jiaqing Xie\textsuperscript{2}},
 \textbf{Yi Ban\textsuperscript{1,2}},
\\
 \textbf{Ben Gao\textsuperscript{2}},
 \textbf{Zhongyu Huang\textsuperscript{2}},
 \textbf{Tianfan Fu\textsuperscript{1,2,*}},
 \textbf{Yuqiang Li\textsuperscript{2,*}}
\\
\\
 \textsuperscript{1}Nanjing University,
 \textsuperscript{2}Shanghai Artificial Intelligence Laboratory
\\
 \small{\textsuperscript{*}Corresponding authors}
}

\begin{document}
\maketitle

\begin{abstract}
Parallel decoding speeds up masked diffusion language models by writing several
tokens per forward pass. Published commit rules choose these tokens by testing
the predicted distributions. Under sampling the written tokens are drawn, and
these tests do not see them. We find that answer quality follows the
reliability of the tokens actually written. An answer's log probability splits
into this reliability and the dependence within each step. We propose PACT, a
training-free commit rule that draws the tokens first, certifies the set it
will write and spaces the positions of each step apart. This
certificate holds at every temperature, and we show that no rule with the same guarantee commits more positions. Experiments use four models of the LLaDA and Dream families and HumanEval, MBPP, GSM8K, MATH500 and IFEval benchmarks. PACT reaches the quality of published
commit rules with fewer forward passes. Spacing removes positions from each
step, yet it usually lowers the total cost because later steps then write more
tokens. At the same speed, published decoders also write a word twice in a row
far more often than PACT. These results suggest that commit rules should certify
the tokens they write. Code is available at
\url{https://github.com/LiZaiyuan0619/PACT}.

\end{abstract}

\section{Introduction}
\label{sec:intro}

\begin{figure}[t]
\centering
\includegraphics[width=\columnwidth]{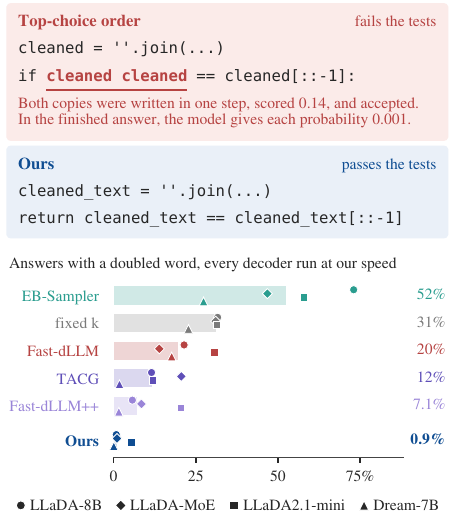}
\caption{\textbf{A test that cannot see the draw writes the same word twice.}
Top, two answers of LLaDA-8B to HumanEval/48 at $\tau = 0.1$. Bottom, the share
of answers with one word written twice in a row at matched forward passes, with
bars at the median over models (Appendix~\ref{app:measures}).}
\label{fig:hero}
\end{figure}

Masked diffusion language models generate text from a fully masked answer
\citep{NEURIPS2024_eb0b13cc, shi2024simplified, nie2025large}. Over a series of
steps, they fill in its positions several at a time. A commit rule decides which
positions each step fills. Filling more of them saves forward passes but raises
the risk of writing a wrong token. Training-free commit rules have moved from a
fixed number of positions per step to a test that sets this number. They make
decoding several times faster \citep{wu2026fastdllm, NEURIPS2025_510e22f4,
kim2025klass, kasa2026fastdllmfrechetprofiledecoding,
wang2026tacgtrajectoryawarecommitgating}. These models are often decoded by
sampling. Reinforcement learning and test-time scaling rely on it for diverse
answers \citep{olausson2026a, fang2026locallyconfidentgloballystuck}.

However, commit rules face two problems when the decoder samples. First,
their tests read the predicted distributions and not the drawn tokens. Under
sampling a drawn token can be far less likely than the most likely one, and the
test misses this gap. Second, the tokens of one step depend on each other, and
this dependence cannot be measured from the distributions of a single forward
pass. In Figure~\ref{fig:hero}, a rule that orders and scores positions by
their top choice accepts \texttt{cleaned} twice in one step, yet the finished
answer gives each copy almost no probability.

We find that quality follows the actual reliability of what a decoder writes,
the log probability of the written tokens. Across rule families and four
models, it predicts pass@1 better than the reliability reported on the most
likely tokens, and far better than the number of forward passes. The log probability of an
answer is also the sum of two terms, the actual reliability of each step and
the dependence among the tokens it writes. This suggests that a rule should
certify the tokens it writes and keep the tokens of one step apart.

We propose PACT, a Post-draw Accumulated Certificate for Tokens. At each step PACT draws the tokens first and orders the positions
by the log probability of their drawn tokens. It then commits the longest prefix
of this order whose log probabilities add up to at least $\log \tau$. The set it
certifies is therefore the set it writes, at any temperature. We prove that it
commits as many positions as this guarantee allows. PACT also spaces the
positions of one step at least a minimum gap apart, to lower the dependence
among them. It is training-free, adds no forward passes and changes only the
commit rule.

We evaluate PACT on four masked diffusion language models from the LLaDA and
Dream families, on HumanEval, MBPP, GSM8K, MATH500 and IFEval. At matched
quality on code and math, it needs fewer forward passes than each of the five
published rules, and the margin grows as decoding gets faster. These savings also shorten the
decoding time. As the temperature
rises, ordering by the top choice collapses, while PACT keeps most of its
accuracy. The doubled word of Figure~\ref{fig:hero} turns up far more often
under the published decoders than under PACT at the same speed. These results
make PACT a faster and more reliable way to decode masked diffusion
language models by sampling.

The main contributions are summarized as follows.
\begin{enumerate}
\item We identify that published commit rules test the predicted distributions
alone, while the written tokens are drawn. Under sampling, we find that quality
follows the actual reliability of what is written.
\item We propose PACT, a commit rule that certifies the drawn tokens it writes
and spaces the positions of one step, with no training and no extra forward
passes.
\item We prove that PACT commits as many positions as its guarantee allows, at
any temperature. Under a Markov assumption, spacing
bounds the dependence within a step.
\item We show on four models that PACT needs fewer forward passes than five
published rules at matched quality.
\end{enumerate}

\begin{figure*}[t]
\centering
\includegraphics[width=\textwidth]{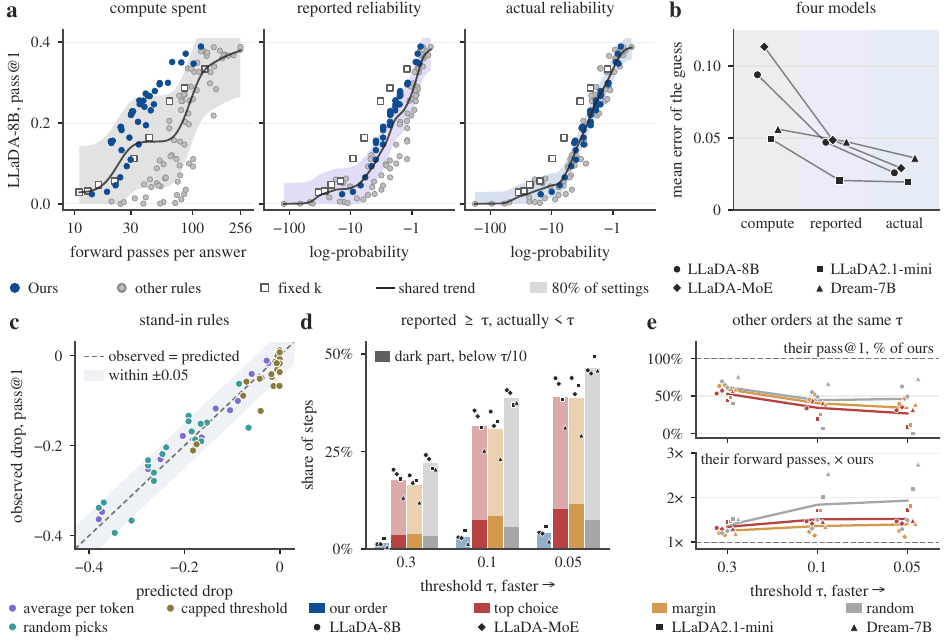}
\caption{\textbf{Quality follows the reliability of what is written, and tests
on the distributions misjudge it.} (a) Pass@1 of LLaDA-8B against three axes.
(b) Error of predicting pass@1 from each axis, holding out one rule family at a
time. (c) Stand-in rules, built to hold another quantity fixed. (d) Steps where the
reported reliability reaches $\tau$ and the actual reliability does not, under
four orders with one shared test. (e) The other three orders against ours.
Appendix~\ref{app:measures} defines each quantity.}
\label{fig:diagnosis}
\end{figure*}

\section{Related Work}
\label{sec:related}

Caching, block-wise decoding, distillation and early exit make masked diffusion
language models faster by changing the cost of each step or the number of steps
(Appendix~\ref{app:related}). They can be combined with the commit rule, the
part of the decoder that picks which masked positions to fill at each step.
Early rules rank the positions by confidence and fill a fixed number per step
\citep{nie2025large}. Adaptive rules let a test decide this number. Some test
each position on its own, by its top probability \citep{wu2026fastdllm,
yu2026dimplediscretediffusionmultimodal}, the stability of its prediction across
steps \citep{kim2025klass} or signals from the decoding trajectory
\citep{wang2026tacgtrajectoryawarecommitgating}. Others sort the positions and
test a prefix as a whole, bounding its summed entropy
\citep{NEURIPS2025_510e22f4} or the joint probability of its most likely tokens
\citep{kasa2026fastdllmfrechetprofiledecoding}. These tests are computed from
the predicted distributions alone, so a step fills the same positions whatever
tokens are drawn. PACT also tests a prefix as a whole. It tests the drawn
tokens, so its certificate covers the set it writes at every temperature
(Section~\ref{sec:pact}).

A step that commits several positions draws each token from its own marginal,
so the dependence between the tokens is left out. Analyses show that the
resulting error lowers quality and moves samples away from the model's
distribution \citep{zhong-etal-2026-parallelism, tang2026is,
zhang2026generationorderparalleldecoding, kang2026parallelbench}.
Dependence-aware rules choose which positions to commit together, from attention
\citep{kim2026dapd}, a trained head
\citep{ringel2026dependencyguidedparalleldecodingdiscrete}, a small
autoregressive model \citep{israel2025accelerating} or the uncertainty of earlier
context \citep{he2026reliableparalleldecodingmasked}. Spreading the positions of
a step apart also weakens their dependence \citep{luxembourg2026plan,
besnier2025halton}. PACT applies this idea inside its certificate, where
positions that lie too close give way to farther ones. The certificate then
bounds one of the two sums that make up the answer's log probability, and
spacing shrinks the other (Section~\ref{sec:theory}). Section~\ref{sec:ablations}
measures what spacing adds to the certificate.

\begin{table*}[t]
\centering
\small
\renewcommand{\arraystretch}{0.88}
\begin{tabular*}{\textwidth}{@{\extracolsep{\fill}}lrrrrrrrrrr}
\toprule
 & \multicolumn{2}{c}{HumanEval} & \multicolumn{2}{c}{MBPP} & \multicolumn{2}{c}{GSM8K} & \multicolumn{2}{c}{MATH500} & \multicolumn{2}{c}{IFEval} \\
\cmidrule(lr){2-3}\cmidrule(lr){4-5}\cmidrule(lr){6-7}\cmidrule(lr){8-9}\cmidrule(lr){10-11}
Method & Acc. & Speed & Acc. & Speed & Acc. & Speed & Acc. & Speed & Acc. & Speed \\
\midrule
\multicolumn{11}{l}{\emph{LLaDA-8B}} \\
One token per step & 38.3 & 1.0$\times$ & 39.1 & 1.0$\times$ & 57.6 & 1.0$\times$ & 27.5 & 1.0$\times$ & 46.1 & 1.0$\times$ \\
Fast-dLLM & \underline{34.8} & 2.9$\times$ & 34.9 & 2.1$\times$ & 57.5 & 2.7$\times$ & 26.0 & 2.0$\times$ & 49.0 & 1.9$\times$ \\
Fast-dLLM++ & 23.2 & \textbf{3.9$\times$} & 23.5 & \textbf{3.0$\times$} & 52.5 & \underline{4.5$\times$} & 24.5 & \textbf{2.7$\times$} & 50.7 & \underline{2.1$\times$} \\
EB-Sampler & 33.2 & \underline{3.3$\times$} & 33.7 & \underline{2.6$\times$} & 57.7 & 3.3$\times$ & 25.9 & 2.3$\times$ & 47.8 & 1.9$\times$ \\
KLASS & 34.4 & 1.4$\times$ & \underline{35.4} & 1.2$\times$ & \underline{58.1} & 1.3$\times$ & 25.8 & 1.1$\times$ & \textbf{52.5} & 1.3$\times$ \\
TACG & 34.7 & 2.8$\times$ & 33.7 & 2.1$\times$ & 57.4 & 2.9$\times$ & \underline{26.1} & 2.0$\times$ & 50.6 & 1.7$\times$ \\
Fixed $k$ ($k{=}4$) & 24.5 & 4.0$\times$ & 25.9 & 4.0$\times$ & 54.9 & 4.0$\times$ & 24.1 & 4.0$\times$ & 42.2 & 4.0$\times$ \\
\textbf{Ours} & \textbf{36.0} & \underline{3.3$\times$} & \textbf{37.0} & 2.2$\times$ & \textbf{59.3} & \textbf{4.7$\times$} & \textbf{27.5} & \underline{2.4$\times$} & \underline{51.9} & \textbf{2.2$\times$} \\
\midrule
\multicolumn{11}{l}{\emph{LLaDA-MoE}} \\
One token per step & 52.1 & 1.0$\times$ & 51.6 & 1.0$\times$ & 44.9 & 1.0$\times$ & 26.8 & 1.0$\times$ & 56.9 & 1.0$\times$ \\
Fast-dLLM & 47.9 & 4.0$\times$ & \underline{47.7} & 2.5$\times$ & \underline{44.1} & 2.2$\times$ & \underline{26.2} & 2.4$\times$ & 58.2 & 1.7$\times$ \\
Fast-dLLM++ & 33.9 & \textbf{5.9$\times$} & 31.2 & \textbf{3.5$\times$} & 39.7 & \textbf{3.7$\times$} & 23.8 & \underline{3.2$\times$} & 57.3 & \textbf{1.9$\times$} \\
EB-Sampler & 48.4 & 4.8$\times$ & 46.6 & \underline{3.0$\times$} & 41.0 & 2.6$\times$ & 25.2 & 2.4$\times$ & \underline{59.3} & 1.7$\times$ \\
KLASS & \underline{49.5} & 1.8$\times$ & 44.6 & 1.2$\times$ & 43.7 & 1.1$\times$ & 24.6 & 1.1$\times$ & 56.2 & 1.2$\times$ \\
TACG & 45.8 & 4.2$\times$ & 45.1 & 2.4$\times$ & 40.4 & 2.5$\times$ & 25.0 & 2.6$\times$ & 59.1 & 1.6$\times$ \\
Fixed $k$ ($k{=}4$) & 38.3 & 4.0$\times$ & 32.6 & 4.0$\times$ & 35.8 & 4.0$\times$ & 23.8 & 4.0$\times$ & 49.4 & 4.0$\times$ \\
\textbf{Ours} & \textbf{50.1} & \underline{5.2$\times$} & \textbf{48.5} & 2.9$\times$ & \textbf{44.9} & \underline{3.5$\times$} & \textbf{26.4} & \textbf{3.7$\times$} & \textbf{60.1} & \underline{1.8$\times$} \\
\midrule
\multicolumn{11}{l}{\emph{LLaDA2.1-mini}} \\
One token per step & 21.5 & 1.0$\times$ & 14.5 & 1.0$\times$ & 28.6 & 1.0$\times$ & 15.7 & 1.0$\times$ & 14.8 & 1.0$\times$ \\
Fast-dLLM & \underline{19.6} & 1.4$\times$ & \underline{13.6} & 1.3$\times$ & 29.2 & \underline{1.5$\times$} & \underline{16.4} & \underline{1.9$\times$} & 14.2 & 1.2$\times$ \\
Fast-dLLM++ & 11.9 & \underline{1.5$\times$} & 9.6 & \underline{1.4$\times$} & 27.7 & 1.4$\times$ & 14.1 & 1.8$\times$ & \underline{19.5} & \underline{1.6$\times$} \\
EB-Sampler & 16.1 & 1.3$\times$ & 12.1 & 1.3$\times$ & \underline{31.6} & 1.3$\times$ & 15.5 & 1.8$\times$ & \textbf{19.6} & 1.3$\times$ \\
KLASS & 16.8 & 1.0$\times$ & 12.6 & 1.0$\times$ & \textbf{33.1} & 1.0$\times$ & 15.8 & 1.1$\times$ & 19.0 & 1.0$\times$ \\
TACG & 17.5 & 1.3$\times$ & 12.2 & 1.2$\times$ & \underline{31.6} & 1.3$\times$ & 15.2 & 1.8$\times$ & 18.2 & 1.3$\times$ \\
Fixed $k$ ($k{=}4$) & 5.6 & 4.0$\times$ & 1.7 & 4.0$\times$ & 14.3 & 4.0$\times$ & 9.7 & 4.0$\times$ & 12.1 & 4.0$\times$ \\
\textbf{Ours} & \textbf{19.8} & \textbf{1.9$\times$} & \textbf{14.8} & \textbf{1.8$\times$} & \underline{31.6} & \textbf{2.3$\times$} & \textbf{16.7} & \textbf{2.9$\times$} & \underline{19.5} & \textbf{1.7$\times$} \\
\midrule
\multicolumn{11}{l}{\emph{Dream-7B}} \\
One token per step & 20.9 & 1.0$\times$ & 25.0 & 1.0$\times$ & 42.3 & 1.0$\times$ & 24.9 & 1.0$\times$ & 26.0 & 1.0$\times$ \\
Fast-dLLM & 22.7 & 1.9$\times$ & 25.2 & 2.2$\times$ & 41.8 & 1.9$\times$ & 25.5 & 1.4$\times$ & 31.0 & 2.7$\times$ \\
Fast-dLLM++ & 16.8 & \underline{3.7$\times$} & 15.7 & \underline{4.7$\times$} & 32.4 & \textbf{4.2$\times$} & 20.9 & \underline{3.3$\times$} & 39.1 & \underline{4.1$\times$} \\
EB-Sampler & \underline{28.3} & 2.0$\times$ & 24.9 & 2.2$\times$ & 41.7 & 2.2$\times$ & \underline{26.7} & 1.4$\times$ & 30.5 & 2.5$\times$ \\
KLASS & 24.4 & 1.2$\times$ & \textbf{30.6} & 1.2$\times$ & \textbf{44.5} & 1.3$\times$ & \textbf{26.8} & 1.1$\times$ & \textbf{42.9} & 1.7$\times$ \\
TACG & 25.5 & 2.3$\times$ & 24.3 & 2.7$\times$ & 39.6 & 2.4$\times$ & 24.9 & 1.8$\times$ & 40.0 & 2.8$\times$ \\
Fixed $k$ ($k{=}4$) & 17.8 & 4.0$\times$ & 14.4 & 4.0$\times$ & 51.7 & 4.0$\times$ & 27.1 & 4.0$\times$ & 37.5 & 4.0$\times$ \\
\textbf{Ours} & \textbf{31.9} & \textbf{5.1$\times$} & \underline{26.2} & \textbf{7.9$\times$} & \underline{43.1} & \underline{3.7$\times$} & 26.2 & \textbf{6.7$\times$} & \underline{41.5} & \textbf{6.7$\times$} \\
\bottomrule
\end{tabular*}
\caption{\textbf{Accuracy and speed of every method at one setting.} Acc. is pass@1 (\%) and Speed is the tokens written per forward pass. Each baseline runs at the setting given in Appendix~\ref{app:baselines}, and PACT at $\tau = 0.5$, without spacing on LLaDA2.1-mini. In each column, the best of the published rules and PACT is in bold and the second best underlined.}
\label{tab:main}
\end{table*}

\section{Method}
\label{sec:method}

\subsection{Preliminaries}
\label{sec:prelim}

A masked diffusion language model starts from a fully masked answer and fills in
part of it at every step \citep{NEURIPS2024_eb0b13cc, shi2024simplified}. At
each step it gives every masked position $i \in M$ a distribution $p_i$ over the
vocabulary, and the decoder draws a token $x_i$ from $q_i \propto p_i^{1/T}$ at
temperature $T$. A commit rule keeps the drawn tokens of a set $S \subseteq M$
and leaves the other positions masked. Each step costs one forward pass.
Committed tokens are not revised.

We describe a commit rule by three parts, a score for each position, an order
over the positions and a cut that ends the ordered list. The cut sets how many
positions a step commits, and the score and the order set which. Published
rules compute their test from the distributions alone, for each position or
summed along the order (Section~\ref{sec:related}). Their decision is therefore
the same whichever tokens are drawn, and it speaks for the written tokens only
under greedy decoding. To make this precise, we compare two quantities for a
set $S$,
\begin{equation}
\label{eq:reliability}
  \hat{R}(S) = \sum_{i \in S} \log p_i(\hat{x}_i),
  \quad
  R(S) = \sum_{i \in S} \log p_i(x_i),
\end{equation}
where $\hat{x}_i = \arg\max_v p_i(v)$ is the most likely token at position $i$.
The reported reliability $\hat{R}(S)$ is computed on the most likely tokens, and
rules that threshold the top probability certify a bound on it. The actual
reliability $R(S)$ is computed on the written tokens and is known only once
they are drawn.

\subsection{PACT}
\label{sec:pact}

\begin{figure*}[t]
\centering
\includegraphics[width=\textwidth]{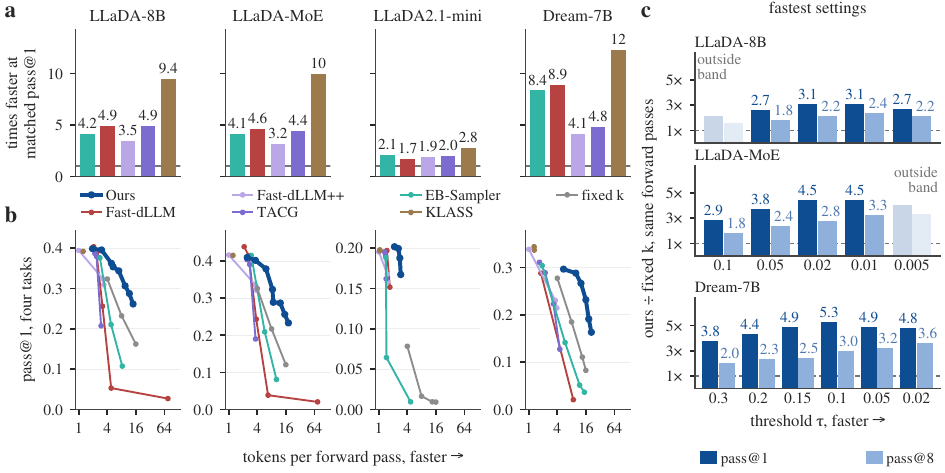}
\caption{\textbf{Speed and quality of every decoder.} (a) Speed-up of PACT over
each published rule at matched pass@1 on code and math. (b) The sweeps behind a.
(c) PACT against fixed k at the same forward passes, with dark bars
significantly above one. Appendix~\ref{app:measures} gives the matched level,
the tasks, the band of c and the test.}
\label{fig:frontier}
\end{figure*}

PACT keeps these three parts and changes each of them. As its name says, it
draws the tokens first and certifies the set it is about to write. Its score
for a position is the written-word score $s_i = \log p_i(x_i)$, the log
probability of the token drawn there. Sorting the positions by this score gives
our order $\pi = (\pi_1, \pi_2, \dots)$. PACT commits the longest prefix
$S = \{\pi_1, \dots, \pi_{k^{\ast}}\}$ of $\pi$ whose scores add up to at least
$\log \tau$, where
\begin{equation}
\label{eq:cut}
  k^{\ast} = \max\Bigl\{\, m \;\Bigm|\; \textstyle\sum_{j \le m} s_{\pi_j} \ge \log \tau \Bigr\}.
\end{equation}
When no prefix passes, it commits $\pi_1$ so that every step makes
progress. This set-level cut tests the committed set as a whole. Because the test reads the drawn tokens, the set PACT certifies is the set it
writes, at any temperature. No rule with this guarantee commits more positions
(Proposition~\ref{prop:pact}).

The set-level cut accounts for how reliable each committed token is, but not for
how the tokens committed together depend on each other. Neighboring positions
are the riskiest pair to commit in one step, since each is decided from a
context that lacks the other \citep{luxembourg2026plan}. The third change
therefore spaces the committed positions apart by filtering the order. Spacing
scans $\pi$ and keeps a position only if it lies at least $g$ away from every
position kept so far. The scan stops once it holds $k^{\ast}$ positions, the
count that \eqref{eq:cut} allows. If too few positions qualify, the scan lowers
the gap one step at a time. Listing the kept positions in the order of $\pi$ as
$\pi^{g}$, PACT commits the longest prefix of $\pi^{g}$ with at most $k^{\ast}$
positions whose scores add up to at least $\log \tau$. The cut is recomputed on $\pi^{g}$, since a position with a lower score can push
the sum below $\log \tau$. Spacing only filters the order, so it adds no forward passes.

Algorithm~\ref{alg:pact} in Appendix~\ref{app:algorithm} lists one step of
PACT. The threshold $\tau$ sets how much reliability each step keeps, and with
it the balance between speed and quality. We set the minimum gap to $g = 8$ by
default. PACT is training-free and changes only the step that picks the
committed set.

\subsection{Theoretical Analysis}
\label{sec:theory}

PACT certifies each step through the tokens it writes. To see what this
controls over a whole answer, let a decode commit $S_1, \dots, S_K$ in this
order, with $R(S_k)$ taken at the state where $S_k$ is committed. For a
committed set $S$, let $P_S(x_S)$ be the probability the model assigns to its
tokens when they are revealed one at a time in the order of $\pi$. Its
dependence term is $D(S) = \log P_S(x_S) - R(S)$.

\begin{proposition}[Decomposition]
\label{prop:decomp}
Let $\log P(x)$ be the log probability the model assigns to the answer when its
tokens are revealed one at a time in commit order. Then
\begin{equation}
\label{eq:decomp}
  \log P(x) = \sum_{k=1}^{K} R(S_k) + \sum_{k=1}^{K} D(S_k).
\end{equation}
If the sets and the order within each set are fixed in advance and \mbox{$T = 1$}, parallel decoding draws the answer
from $Q(x) = \prod_{k} \prod_{i \in S_k} p_i(x_i)$, and
$\mathbb{E}_{Q}[\sum_{k} D(S_k)] = -\mathrm{KL}(Q \,\|\, P) \le 0$.
\end{proposition}

\begin{figure*}[t]
\centering
\includegraphics[width=\textwidth]{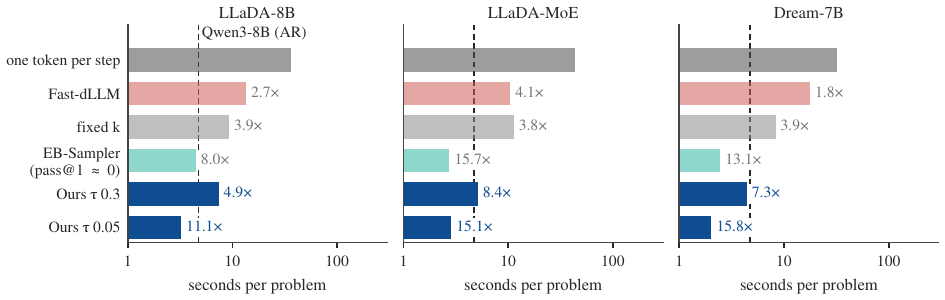}
\caption{\textbf{Decoding time per problem.} Multiples are speed-ups over one
token per step (Appendix~\ref{app:cost}).}
\label{fig:speed}
\end{figure*}

The first sum collects the reliability of what each step writes, and a rule can
check it once the tokens are drawn. The second sum is the cost of writing
several tokens at once, and a single forward pass does not reveal it. The threshold $\tau$ bounds every term of the first sum,
and spacing shrinks the terms of the second. The first sum tracks
quality. Across rule families and four models, pass@1 follows the actual
reliability of the weakest steps, while the reported reliability
predicts it worse and compute far worse (Figure~\ref{fig:diagnosis}a--b). Rules built to hold
another quantity fixed lose about the pass@1 that this relation predicts
(Figure~\ref{fig:diagnosis}c).

\begin{proposition}[Distribution-only tests]
\label{prop:blind}
For every set $S$, $\hat{R}(S) \ge R(S)$, with equality if and only if every
position in $S$ receives its most likely token. If $S$ is chosen from the
distributions alone, $\mathbb{E}_T[R(S)]$ is non-increasing in $T$ and tends to
$\hat{R}(S)$ as $T \to 0$ when each most likely token is unique. At $T = 1$,
the expected gap $\mathbb{E}[\hat{R}(S) - R(S)]$ is
$\sum_{i \in S} [H(p_i) - H_{\infty}(p_i)]$, where $H$ and $H_{\infty}$ are the
entropy and the min-entropy.
\end{proposition}

A rule that decides from the distributions alone commits the same set at a
given state whatever the temperature. The tokens it writes there lose
reliability as the temperature rises. The gap closes only under greedy
decoding, the setting in which guarantees on the most likely tokens apply.
Within single steps, orders that ignore the draw often report a reliability at
or above $\tau$ while the written tokens fall below it
(Figure~\ref{fig:diagnosis}d). Ordering by the drawn tokens rarely leaves such a gap, and at the same $\tau$ it
reaches a higher pass@1 with fewer forward passes
(Figure~\ref{fig:diagnosis}e, Appendix~\ref{app:orders}).

\begin{proposition}[Largest certified set]
\label{prop:pact}
Except when no prefix passes and a single position is committed, every set PACT
commits satisfies $R(S) \ge \log \tau$, at every temperature and after any
filter that only removes positions from $\pi$. Without spacing, PACT commits the
largest set $S$ with $R(S) \ge \log \tau$. As $\tau \to 1$, PACT commits one
position per step, and with $g = 1$ spacing leaves $\pi$ unchanged.
\end{proposition}

The set PACT certifies is thus the set it writes, and spacing keeps it so. The
guarantee also costs no parallelism within a step, since any rule with the same
guarantee commits at most as many positions. Decoding one token per step is the
end of this family where $\tau$ approaches one.

\begin{figure*}[t]
\centering
\includegraphics[width=\textwidth]{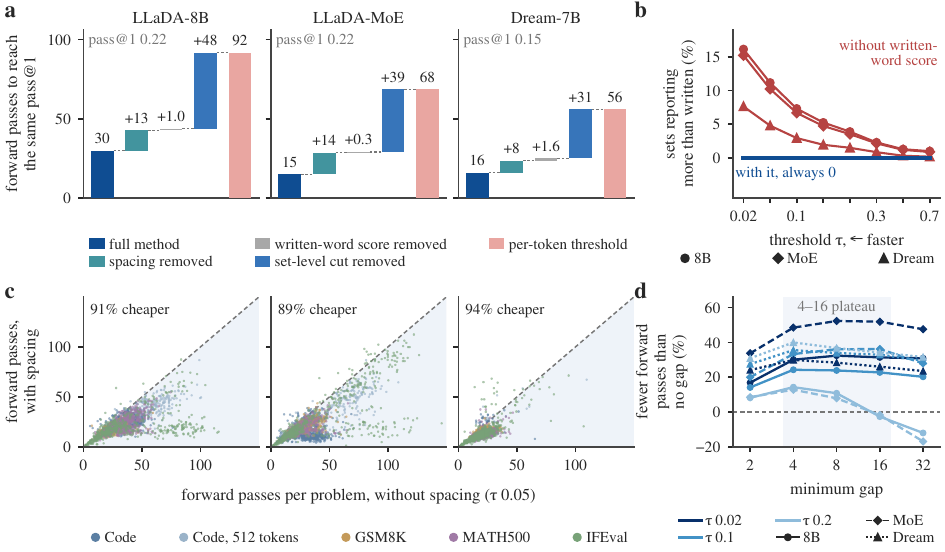}
\caption{\textbf{Ablations of PACT.} (a) Forward passes on code to
reach the pass@1 printed in each panel, as the parts of PACT are removed in
turn. (b) Committed sets on code that report more than they write. (c) Forward
passes of each problem with and without spacing, on the models of a. (d) The
saving of spacing on code by minimum gap.}
\label{fig:ablation}
\end{figure*}

\begin{proposition}[Spacing bounds the dependence term]
\label{prop:gap}
Suppose that at a given state the model's conditionals come from one joint
distribution $P$ over the masked positions, with marginals $p_i$, and that $P$
is Markov along the sequence. Let the $m$ positions of $S$ lie pairwise at least
$g$ apart, and let $J_I(g)$ and $J_L(g)$ be the largest mutual information and
the largest lautum information $L(X;Y) = \mathrm{KL}(P_X P_Y \,\|\, P_{XY})$
between two positions $g$ apart. Then
\[
\begin{aligned}
  0 &\le \mathbb{E}_{P_S}\bigl[D(S)\bigr] \le (m-1)\, J_I(g), \\
  0 &\le -\mathbb{E}_{\otimes_i p_i}\bigl[D(S)\bigr] \le (m-1)\, J_L(g),
\end{aligned}
\]
and both $J_I$ and $J_L$ are non-increasing in $g$.
\end{proposition}

The first expectation draws the set from the joint, and the second draws each
position on its own, as parallel decoding does at $T = 1$. Either way, the sum of marginals
misjudges the set by at most $m - 1$ times the dependence between positions $g$
apart. This dependence shrinks as $g$ grows. The factor $m - 1$ makes spacing matter more
when a step writes many tokens. Both assumptions are idealizations, so Section~\ref{sec:analysis} tests only the trend they predict.
Appendix~\ref{app:proofs} gives the proofs.

\section{Experiments}
\label{sec:experiment}

\subsection{Setup}
\label{sec:setup}

\paragraph{Models and benchmarks.}
We evaluate PACT on four masked diffusion language models, LLaDA-8B, LLaDA-MoE,
LLaDA2.1-mini and Dream-7B. Two are dense and two are mixtures of experts, from
the LLaDA and Dream families. We use HumanEval and MBPP for code, GSM8K and
MATH500 for math, and IFEval for instruction following
(Appendices~\ref{app:models} and~\ref{app:tasks}).

\paragraph{Baselines.}
We compare PACT with five published commit rules, Fast-dLLM (a per-token threshold), Fast-dLLM++,
EB-Sampler, KLASS and TACG. All of them follow the definitions in their papers and
run in our decoder under the same protocol as PACT (Appendix~\ref{app:baselines}). Two simple
decoders serve as references, one token per step and fixed k. Fixed k commits
at every step the $k$ positions whose drawn tokens have the highest probability. For decoding time, we add Qwen3-8B, an autoregressive model of similar size.

\paragraph{Metrics and protocol.}
We report pass@1 over eight samples per problem, and pass@8 where noted. Cost is
the number of forward passes per answer. Each answer fills all 256 positions,
including the padding after it ends. The tokens written per forward pass
therefore equal the speed-up over one token per step. Each problem is decoded at
temperature 0.6 as a single block of 256 tokens, so the commit rule alone decides
the decoding order. The diffusion models run without a key-value cache (Appendix~\ref{app:cost}). PACT uses $g = 8$ except on LLaDA2.1-mini, which runs without spacing
(Section~\ref{sec:analysis}). Differences are tested with a paired bootstrap, and Appendix~\ref{app:stats} reports the variation across seeds. Appendix~\ref{app:blocks} repeats the comparison with
32-token blocks.

\subsection{Main Results}
\label{sec:main}

\textbf{At one setting each, published rules that are more accurate than PACT
are almost always slower (Table~\ref{tab:main}).} We first run each
method at one setting, as a user would. The per-token threshold is slower
than PACT on every model and benchmark, at similar accuracy. Since the methods
differ in speed and accuracy at once, Figure~\ref{fig:frontier} compares them at
matched quality and at matched speed.

\begin{figure*}[t]
\centering
\includegraphics[width=\textwidth]{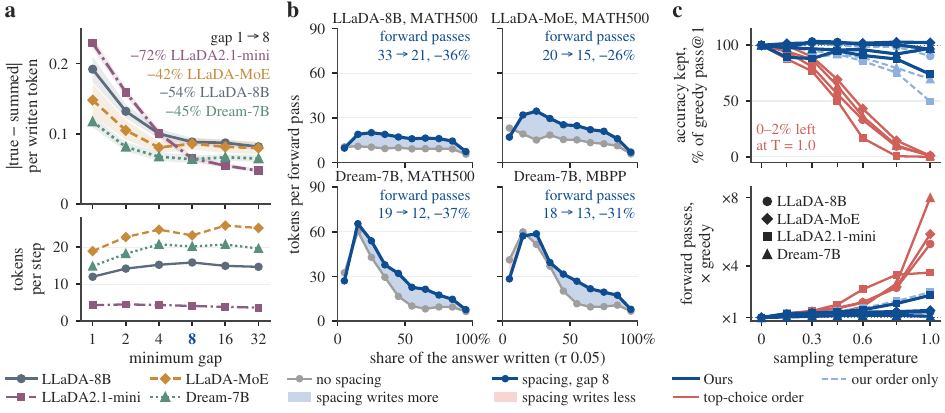}
\caption{\textbf{Effects of spacing and of the sampling
temperature.} (a) Dependence term and step size against the minimum gap. (b)
Tokens written per forward pass along the answer. (c) Accuracy and forward
passes against the sampling temperature, for three rules that share the
set-level cut. Appendix~\ref{app:measures} gives the settings.}
\label{fig:analysis}
\end{figure*}

\textbf{At matched quality, PACT needs fewer forward passes than every published
rule on every model (Figure~\ref{fig:frontier}a--b).} Figure~\ref{fig:frontier}a
compares the fastest setting of each method that reaches the same pass@1. PACT
is close to the published rules when each forward pass writes few tokens, and
it pulls ahead as decoding gets faster (Figure~\ref{fig:frontier}b).

\textbf{At the fastest settings, PACT is several times more accurate than fixed
k at the same forward passes (Figure~\ref{fig:frontier}c).}
Figure~\ref{fig:frontier}c pairs each fast setting of PACT with fixed k. PACT
leads on pass@8 as well as on pass@1, and its answers stay as varied as those
of one token per step (Appendix~\ref{app:notes}). Unlike fixed k, PACT lets the reliability of each step decide
how many tokens it writes, more where the answer is easy and fewer where it is
hard.

\textbf{The forward passes PACT saves carry over to decoding time
(Figure~\ref{fig:speed}).} PACT adds only sorting and summing to each step, so
the time it saves tracks the forward passes it saves. Matched to the accuracy of
one token per step, PACT is faster on almost every benchmark
(Appendix~\ref{app:wallclock}). Qwen3-8B decodes eight answers almost as fast as
one with its key-value cache. With one answer per problem, PACT is faster than
it at both thresholds (Appendix~\ref{app:ar}).

\subsection{Ablations}
\label{sec:ablations}

\textbf{At the same pass@1, the set-level cut accounts for most of the forward
passes PACT saves over the per-token threshold (Figure~\ref{fig:ablation}a).}
Figure~\ref{fig:ablation}a removes the parts of PACT one at a time. This matches
Proposition~\ref{prop:pact}, by which the set-level cut commits the largest set
that meets its guarantee. Spacing adds a further saving.

\textbf{The written-word score makes every committed set report the reliability
of exactly what it writes (Figure~\ref{fig:ablation}b).} Without it, a committed
set can report a higher reliability than the tokens it writes
(Appendix~\ref{app:notes}). With it, the two reliabilities are one quantity, and
Proposition~\ref{prop:pact} turns the threshold into a guarantee on what is
written. This costs almost no forward passes (Figure~\ref{fig:ablation}a), and
its effect on accuracy grows with the sampling temperature
(Section~\ref{sec:analysis}).

\textbf{Spacing saves forward passes on almost every problem, and the saving
grows as decoding gets faster (Figure~\ref{fig:ablation}c--d).}
Figure~\ref{fig:ablation}c compares the forward passes of each problem with and
without spacing, and Figure~\ref{fig:ablation}d varies the minimum gap. The
saving grows as $\tau$ falls and each step writes more tokens, as the factor
$m - 1$ in Proposition~\ref{prop:gap} suggests. At the faster thresholds, gaps
on either side of $g = 8$ save about as much, so the choice of $g$ is not
critical.

\subsection{Analysis}
\label{sec:analysis}

\textbf{Spacing shrinks the dependence term by about half
(Figure~\ref{fig:analysis}a).} The term per written token falls as the gap
widens, the trend Proposition~\ref{prop:gap} predicts when dependence decays
with distance. On three of the models the steps also grow with the gap, so the
reduction does not come from smaller steps.

\textbf{Spacing lets later steps write more tokens, so an answer needs fewer
forward passes (Figure~\ref{fig:analysis}b).} Figure~\ref{fig:analysis}b
follows the tokens written per forward pass along an answer. Within a step,
spacing commits at most $k^{\ast}$ positions, so it can only shorten the
committed prefix. The saving comes from the later steps, which write
more tokens once the committed ones are spread across the answer. On
LLaDA2.1-mini the steps stay near four tokens at every gap
(Figure~\ref{fig:analysis}a), so spacing has little to gain and we run it
without spacing.

\textbf{As the sampling temperature rises, ordering by the top choice loses
accuracy far faster than PACT (Figure~\ref{fig:analysis}c).}
Figure~\ref{fig:analysis}c raises the temperature for three rules that differ
only in how they order and score positions (Appendix~\ref{app:notes}). Ordering
by the drawn tokens avoids the collapse, and scoring them as well keeps more of
the accuracy. GSM8K shows the same pattern
(Appendix~\ref{app:temperature}). Proposition~\ref{prop:blind} predicts this,
since the gap of a rule that decides from the distributions alone widens with
the temperature.

\textbf{At the same speed, the published decoders write doubled words far more
often than PACT (Figure~\ref{fig:hero}).} Two copies of a word can each be
likely on their own and unlikely together. A test built from single positions
cannot see this, and it accepts both in one step. Spacing keeps neighboring
positions out of the same step, so such pairs rarely form under PACT.

\section{Conclusion}
\label{sec:conclusion}

This paper revisits what a commit rule certifies when masked diffusion language
models are decoded by sampling. We showed that published rules decide from the
predicted distributions alone, while answer quality depends on how reliable the
written tokens are. PACT certifies the written tokens and spaces the positions within a step,
acting on the two terms of the answer's log probability. On four models, it saved forward passes at matched quality and wrote
far fewer doubled words at matched speed. Speed-ups in parallel decoding are
usually framed as how much risk each step may take. Our results suggest that
which tokens the risk is measured on matters as well. A natural next step is to
certify the joint probability of what each step writes. In uses that sample many
answers, what a commit rule certifies may matter as much as how many tokens it
commits.

\section*{Limitations}
Like other training-free commit rules, PACT sees one distribution per position
in each forward pass. Its certificate therefore covers the summed log probability
of the written tokens. Their joint probability also contains the dependence
among them, which cannot be measured from these distributions when the step is
decided. Spacing lowers
this dependence indirectly, and bounds on it rest on assumptions about how
dependence decays along the sequence. Measuring it directly would take extra
forward passes or a trained estimator.

The certificate rests on the model's own probabilities. It limits the error
that parallel decoding adds, and it cannot correct a model that is confidently
wrong. The outputs also keep the biases of the underlying model, and they need
the same checks as any model output.

Our experiments cover four masked diffusion language models from two families
and English benchmarks of code, math and instruction following, all scored
automatically. Longer generations, other languages and open-ended tasks judged
by people are left to future work.

We count cost in forward passes. The time that a saved pass is worth depends on
the rest of the inference stack, since caches, block-wise models and batching
all change the cost of a pass.

\bibliography{ref}

\appendix
\section{Related Work in Detail}
\label{app:related}

Several lines of work make masked diffusion language models faster without
changing the commit rule. Caching reuses computation across steps
\citep{NEURIPS2025_db0ee27c}. Block-wise decoding generates the answer block by
block, and later work adapts the block size or the budget of steps
\citep{arriola2025block, shen-etal-2026-improving,
wu2026blockbatchmultiscaleconsensusdecoding,
lu2026adaptivetesttimecomputeallocation}. Distillation trains a model to write
more tokens per step \citep{chen2026dparallel}, and early exit stops decoding
once the answer is settled \citep{li2026diffusion}. These methods change the
cost of a step or the number of steps, and each can be combined with a commit
rule. Sampled answers also matter for test-time scaling, which searches over
many of them \citep{bai2026efficient}.

Another line studies the order and the schedule of decoding. Some work sets how
many positions each step decodes from the error of parallel decoding
\citep{park2025jump, chen2025optimalinferenceschedulesmasked}. Others ask how
much an arbitrary decoding order helps \citep{ni2026the} and how these models
behave across tasks and decoding settings
\citep{bertolani2026diffusionlanguagemodelsexperimental}.

\section{Method Details}
\label{app:method}

\subsection{Algorithm}
\label{app:algorithm}

Algorithm~\ref{alg:pact} gives one step of PACT in the notation of
Section~\ref{sec:pact}. Besides the forward pass of the model, a step draws the
tokens, sorts the masked positions, adds up scores and scans the order. It adds
no forward passes, and Appendix~\ref{app:wallclock} shows that its decoding time
follows its forward passes. The scan keeps the full gap $g$ whenever $k^{\ast}$
positions lie that far apart, and lowers it one step at a time otherwise.
Lowering the gap keeps the certificate, since the scan still keeps a subsequence
of $\pi$ (Proposition~\ref{prop:pact}). Sorting takes $O(|M| \log |M|)$
operations and each scan $O(|M| k^{\ast})$, far fewer than a forward pass.

\begin{algorithm}[t]
\caption{One step of PACT}
\label{alg:pact}
\begin{algorithmic}[1]
\Require masked positions $M$, distributions $p_i$, temperature $T$, threshold
  $\tau$, minimum gap $g$
\For{each $i \in M$}
  \State draw $x_i$ from $q_i \propto p_i^{1/T}$ and set $s_i = \log p_i(x_i)$
\EndFor
\State $\pi \gets$ the positions of $M$ sorted by $s_i$, highest first
\State $k^{\ast} \gets$ the largest $m$ with $\sum_{j \le m} s_{\pi_j} \ge \log \tau$,
  or $1$ if no $m$ passes
\State $\pi^{g} \gets$ empty list, $h \gets g$
\While{$\pi^{g}$ holds fewer than $k^{\ast}$ positions}
  \For{$j = 1, \dots, |M|$}
    \If{$\pi^{g}$ holds $k^{\ast}$ positions}
      \State \textbf{break}
    \EndIf
    \If{$\pi_j \notin \pi^{g}$ and $\pi_j$ lies at least $h$ away from every
      position in $\pi^{g}$}
      \State add $\pi_j$ to $\pi^{g}$
    \EndIf
  \EndFor
  \State $h \gets h - 1$
\EndWhile
\State list $\pi^{g}$ in the order of $\pi$
\State $k \gets$ the largest $m \le k^{\ast}$ with
  $\sum_{j \le m} s_{\pi^{g}_j} \ge \log \tau$, or $1$ if no $m$ passes
\State commit $x_i$ at the positions $\pi^{g}_1, \dots, \pi^{g}_{k}$
\end{algorithmic}
\end{algorithm}

\subsection{Proofs}
\label{app:proofs}

Every statement concerns a single step at a fixed state, except
Proposition~\ref{prop:decomp}, which concerns a whole decode. We write $p_i$ for
the distribution of position $i$ at the state where it is committed.

\paragraph{Proof of Proposition~\ref{prop:decomp}.}
Reveal the answer one token at a time in commit order. By the chain rule, the
model assigns it $P(x) = \prod_{k=1}^{K} P_{S_k}(x_{S_k})$, where $P_{S_k}$ conditions on every token committed before step $k$. By the
definition of $D$, $\log P_{S_k}(x_{S_k}) = R(S_k) + D(S_k)$, and summing over
$k$ gives \eqref{eq:decomp}.

Now fix the sets and the order within each set in advance, and take $T = 1$. The
product $P(x)$ then multiplies the model's conditionals along one fixed order, so
it sums to one over answers. Every position is drawn from its own marginal, so parallel decoding draws the answer with probability
$Q(x) = \prod_{k} \prod_{i \in S_k} p_i(x_i)$, and $\log Q(x) = \sum_{k} R(S_k)$.
Subtracting this from \eqref{eq:decomp} and taking the expectation under $Q$,
\[
\begin{aligned}
  \mathbb{E}_{Q}\Bigl[\textstyle\sum_{k} D(S_k)\Bigr]
  &= \mathbb{E}_{Q}\bigl[\log P(x) - \log Q(x)\bigr] \\
  &= -\mathrm{KL}(Q \,\|\, P) \ \le\ 0 .
\end{aligned}
\]
The sets and the order have to be fixed in advance. An adaptive rule chooses its
sets from the drawn tokens. An order that depends on the drawn tokens can also
keep $P(x)$ from summing to one, unless the conditionals come from one joint
distribution.

\paragraph{Proof of Proposition~\ref{prop:blind}.}
Each term of the difference satisfies
$\log p_i(\hat{x}_i) - \log p_i(x_i) \ge 0$, with equality exactly when $x_i$ is a most likely token. Summing over $S$ gives
the inequality and its equality case.

For the dependence on $T$, fix one position and drop its index. With
$\beta = 1/T$, the token is drawn from
$q_\beta(v) = p(v)^{\beta} / Z(\beta)$, where $Z(\beta) = \sum_{v} p(v)^{\beta}$.
The expected score $m(\beta) = \mathbb{E}_{q_\beta}[\log p(x)]$ is the derivative
of $\log Z$, and as in any exponential family its own derivative is a variance,
\[
\begin{aligned}
  m(\beta) &= \frac{d}{d\beta} \log Z(\beta), \\
  m'(\beta) &= \frac{d^2}{d\beta^2} \log Z(\beta)
  = \mathrm{Var}_{q_\beta}\bigl(\log p(x)\bigr) \ \ge\ 0 .
\end{aligned}
\]
Hence $m$ is non-decreasing in $\beta$ and non-increasing in $T$. As
$\beta \to \infty$ the weights $q_\beta$ concentrate on the most likely token
when it is unique, and $m(\beta) \to \log p(\hat{x})$. At $\beta = 1$ the draw
follows $p$ itself, so $m(1) = -H(p)$ and
$\log p(\hat{x}) - m(1) = \log \max_v p(v) + H(p) = H(p) - H_\infty(p)$, with
$H_\infty(p) = -\log \max_v p(v)$ the min-entropy. A set chosen from the
distributions alone does not depend on the draws at this state, so
$\mathbb{E}_T[R(S)] = \sum_{i \in S} m_i(1/T)$, and each statement follows by summing over $S$.

\paragraph{Proof of Proposition~\ref{prop:pact}.}
When the maximum in \eqref{eq:cut} is over a non-empty set, the committed set
satisfies $R(S) = \sum_{j \le k^{\ast}} s_{\pi_j} \ge \log \tau$ by construction. The scores are computed on the drawn tokens, so the bound holds
whatever distribution the tokens were drawn from, at any temperature and under
top-k or top-p truncation. A filter
that only removes positions yields a subsequence of $\pi$, and the cut
recomputed along it satisfies the same bound by the same
argument.

For the largest set, let $S'$ have $m$ positions and $R(S') \ge \log \tau$. The
$m$ largest scores add up to at least the scores of $S'$,
\[
  \sum_{j \le m} s_{\pi_j} \ \ge\ \sum_{i \in S'} s_i \ =\ R(S') \ \ge\ \log \tau ,
\]
so the prefix of $\pi$ of length $m$ passes. Every score is at most zero, so
prefix sums do not increase with length, and the longest passing prefix has
length $k^{\ast} \ge m$.

For the endpoint, committing two positions requires
$s_{\pi_1} + s_{\pi_2} \ge \log \tau$, that is
$p_{\pi_1}(x_{\pi_1})\, p_{\pi_2}(x_{\pi_2}) \ge \tau$. The product is below one unless both drawn tokens have probability one, so for
$\tau$ close enough to one only $\pi_1$ is committed. With $g = 1$ any two
distinct positions lie at least $g$ apart, so the scan takes the positions of
$\pi$ in order, $\pi^{g}$ is the prefix of $\pi$ of length $k^{\ast}$, and
the recomputed cut returns the set of \eqref{eq:cut}.

\paragraph{Proof of Proposition~\ref{prop:gap}.}
List the positions of $S$ in sequence order as $s_1 < \dots < s_m$, so that
consecutive ones lie at least $g$ apart. Because the conditionals come from one
joint $P$, the probability $P_S$ does not depend on the order of revealing.
Because $P$ is Markov along the sequence, so is the subsequence
$X_{s_1}, \dots, X_{s_m}$.

For the first bound, the expected dependence term under $P_S$ is the total
correlation of $X_S$, the expectation under $P$ of
$\log P(X_S) - \sum_{j} \log p_{s_j}(X_{s_j})$. The chain rule and the Markov
property give
\[
\begin{aligned}
  \mathbb{E}_{P_S}\bigl[D(S)\bigr]
  &= \textstyle\sum_{j \ge 2} I\bigl(X_{s_j}; X_{s_1}, \dots, X_{s_{j-1}}\bigr) \\
  &= \textstyle\sum_{j \ge 2} I\bigl(X_{s_j}; X_{s_{j-1}}\bigr) \ \ge\ 0 .
\end{aligned}
\]

For the second bound, the Markov property gives
\[
\begin{aligned}
  \log P(x_S) &= \log p_{s_1}(x_{s_1}) \\
  &\quad + \textstyle\sum_{j \ge 2} \log P\bigl(x_{s_j} \mid x_{s_{j-1}}\bigr).
\end{aligned}
\]
Write $Q_j = p_{s_{j-1}} \otimes p_{s_j}$ for the product of the marginals of a
consecutive pair and $P_j$ for their joint. When every position is drawn from
its own marginal,
\[
\begin{aligned}
  -\mathbb{E}_{\otimes_i p_i}\bigl[D(S)\bigr]
  &= \textstyle\sum_{j \ge 2} \mathbb{E}_{Q_j}\bigl[\log Q_j - \log P_j\bigr] \\
  &= \textstyle\sum_{j \ge 2} L\bigl(X_{s_{j-1}}; X_{s_j}\bigr) \ \ge\ 0 .
\end{aligned}
\]

It remains to bound one pair. Let $a$ and $a + d$ be two positions with
$d \ge g$. Since $X_a \to X_{a+g} \to X_{a+d}$ is a Markov chain, the data
processing inequality gives
\[
  I(X_a; X_{a+d}) \le I(X_a; X_{a+g}) \le J_I(g).
\]
For the lautum information, pass the second coordinate of both
$P_{X_a} P_{X_{a+g}}$ and $P_{X_a X_{a+g}}$ through the channel
$P(X_{a+d} \mid X_{a+g})$. The first becomes $P_{X_a} P_{X_{a+d}}$, and by the
Markov property the second becomes $P_{X_a X_{a+d}}$. A divergence does not
increase under a common channel, so
\[
  L(X_a; X_{a+d}) \le L(X_a; X_{a+g}) \le J_L(g).
\]
Each of the $m - 1$ terms is therefore at most $J_I(g)$ or $J_L(g)$. The same
inequalities with $d = g + 1$ show that $J_I$ and $J_L$ are non-increasing in
$g$.

\section{Experimental Details}
\label{app:setup}

\subsection{Models}
\label{app:models}

Table~\ref{tab:artifacts} lists the checkpoints of LLaDA-8B, LLaDA-MoE
\citep{zhu2025llada}, LLaDA2.1-mini \citep{bie2026llada2}, Dream-7B
\citep{ye2025dream7bdiffusionlarge} and Qwen3-8B \citep{yang2025qwen3}, with
their licenses. Each model decodes with
its own tokenizer and chat template. LLaDA2.1-mini also decodes
the answer as one block and without token editing, so committed tokens stay
fixed for every model.

\begin{table*}[t]
\centering
\small
\begin{tabular}{@{}lll@{}}
\toprule
Checkpoint & Parameters & License \\
\midrule
LLaDA-8B-Instruct         & 8B  & MIT \\
LLaDA-MoE-7B-A1B-Instruct & 7B  & Apache-2.0 \\
LLaDA2.1-mini             & 16B & Apache-2.0 \\
Dream-v0-Instruct-7B      & 7B  & Apache-2.0 \\
Qwen3-8B                  & 8B  & Apache-2.0 \\
\bottomrule
\end{tabular}
\hspace{2em}
\begin{tabular}{@{}lll@{}}
\toprule
Benchmark & Problems used & License \\
\midrule
HumanEval & all 164                    & MIT \\
MBPP      & all 500 test problems      & CC BY 4.0 \\
GSM8K     & 500 of 1,319 test problems & MIT \\
MATH500   & all 500                    & MIT \\
IFEval    & first 500 of 541           & Apache-2.0 \\
\bottomrule
\end{tabular}
\caption{\textbf{Models and benchmarks.} Checkpoints are the releases on Hugging
Face. Every method decodes the same problems of each benchmark.}
\label{tab:artifacts}
\end{table*}

\subsection{Benchmarks}
\label{app:tasks}

Table~\ref{tab:artifacts} also gives the problems we use from each benchmark.
HumanEval \citep{chen2021evaluating} and MATH500 \citep{lightman2024lets} are
used in full, and MBPP \citep{austin2021programsynthesislargelanguage}
contributes its 500 test problems. For GSM8K
\citep{cobbe2021trainingverifierssolvemath} we shuffle the test split once under
a fixed seed and keep the first 500 problems. IFEval
\citep{zhou2023instructionfollowingevaluationlargelanguage} contributes the
first 500 of its 541 prompts. Keeping 500 problems of each bounds the cost of
our sweeps, which all run on the same problems.

The prompt holds the question. For HumanEval and MBPP, a fixed instruction asks
the model to complete the given function. A MATH500 question ends with an
instruction to put the final answer in a box. Our code gives the exact prompts.
We place the prompt in a single user turn and apply
the chat template of the model, with no system message and no examples. Each
answer has 256 positions.

A code answer is scored by running the first code block it contains against the
tests of the benchmark. A math answer is correct when its final answer is
equivalent to the reference. IFEval uses its official strict check, and an
answer counts as correct when it follows every instruction in its prompt.

\subsection{Baselines}
\label{app:baselines}

We implement the five published rules in the same decoder as PACT, following
the equations and algorithms of their papers. Every rule writes the tokens drawn
at the sampling temperature, so the rules differ only in which positions they
commit. Fast-dLLM scores a position by the probability of its most likely token,
as in Algorithm~1 of its paper.

Table~\ref{tab:settings} lists the parameter of each rule, its value in
Table~\ref{tab:main} and its range in Figure~\ref{fig:frontier}.
Table~\ref{tab:main} runs each rule at one value on every model and benchmark.
Each point of Figure~\ref{fig:frontier} is a setting that was run on all four
code and math benchmarks of a model. Not every rule is swept on IFEval, so
Figure~\ref{fig:frontier} leaves it out. KLASS keeps its default thresholds on
LLaDA-8B and LLaDA-MoE.

\begin{table}[t]
\centering
\small
\begin{tabular}{llll}
\toprule
Rule & Parameter & Table~\ref{tab:main} & Figure~\ref{fig:frontier} \\
\midrule
Fast-dLLM   & threshold  & 0.9 & 0.1 to 0.95 \\
Fast-dLLM++ & $\delta$   & 0   & 0 to 1 \\
EB-Sampler  & $\gamma$   & 0.5 & 0.5 to 24 \\
KLASS       & confidence & 0.9 & 0.3 to 0.9 \\
TACG        & threshold  & 0.9 & 0.5 to 0.95 \\
Fixed $k$   & $k$        & 4   & 4 to 16 \\
PACT        & $\tau$     & 0.5 & 0.02 to 0.7 \\
\bottomrule
\end{tabular}
\caption{\textbf{Settings of each rule.} KLASS keeps its KL threshold at 0.001.
PACT uses $g = 8$, without spacing on LLaDA2.1-mini, and
Figure~\ref{fig:frontier} also sweeps it without spacing.}
\label{tab:settings}
\end{table}

\subsection{Cost and Timing}
\label{app:cost}

Every method fills the 256 positions of an answer in the same way, so forward
passes count the same work across methods. On every model and benchmark, PACT
writes answers about as long as those of one token per step, or longer. The cost of a rule includes every
forward pass it makes. The eight answers of a problem are decoded together, so
the cost of each is the number of forward passes until all eight are filled. No diffusion decoder in our comparison uses a key-value
cache, which keeps the comparison to the commit rule itself.

Times are seconds on NVIDIA H200 GPUs, the hardware of every experiment in this
paper. In Figure~\ref{fig:speed} and Table~\ref{tab:ar}, each diffusion model
and Qwen3-8B are timed on the same GPU, and Qwen3-8B decodes with a key-value
cache through Hugging Face Transformers. The times in
Figure~\ref{fig:app_time} are recorded by the runs of our sweeps.

Figure~\ref{fig:speed} times 50 HumanEval problems, with the eight answers of
each problem decoded together as one block of 256 tokens. Fast-dLLM and fixed k
run at their settings in Table~\ref{tab:main}, and EB-Sampler at $\gamma = 16$
on LLaDA-8B and $\gamma = 24$ on the other two models. On the full HumanEval set,
PACT keeps 80\% to 98\% of the pass@1 of one token per step at $\tau = 0.3$.

\subsection{Statistics}
\label{app:stats}

Intervals and tests come from a paired bootstrap over problems. Each resample
draws problems with replacement and keeps the two methods paired on each
problem. We draw 10,000 resamples under a fixed seed and take the 2.5th and
97.5th percentiles as the 95\% interval.

Our runs use seed 113. Rerunning a setting with seed 217 or 331 changes pass@1
by 1.0 point at the median and pass@8 by 2.4 points. These medians are over 24
reruns of Fast-dLLM and PACT on the code benchmarks of LLaDA-8B and LLaDA-MoE.

\subsection{Measurements}
\label{app:measures}

For each committed set $S$ we record the reported reliability $\hat{R}(S)$ and
the actual reliability $R(S)$ of \eqref{eq:reliability}.
Figure~\ref{fig:diagnosis}a summarizes a setting by the 5th percentile of each
over the steps of its answers, which reflects the weakest steps of a decode.

Figure~\ref{fig:diagnosis}b predicts the pass@1 of each rule family from one
axis. We hold out one family, fit a monotone curve of pass@1 against the axis on
the settings of the other families, and read the held-out settings off this
curve. The error is the mean absolute difference between the observed and the
predicted pass@1, and settings outside the range of the curve are not predicted.
The trend line in Figure~\ref{fig:diagnosis}a is the same fit on every setting.

Figure~\ref{fig:diagnosis}c tests this relation with stand-in rules, built to
hold one other quantity fixed. Each of them matches a reference rule in the number of
positions per step, in forward passes or in the average score per token, and
commits other positions. We pair each with a reference setting within 15\% of
its forward passes. Its predicted drop in pass@1 is read from the curve of
Figure~\ref{fig:diagnosis}a at its actual reliability.

Figure~\ref{fig:diagnosis}d and e compare four orders that share one set-level
cut on the reported reliability. The top-choice order ranks positions by the
probability of their most likely token, the margin order by the gap between
their two most likely tokens, and the random order at random. Our order ranks
them by the written-word score. Figure~\ref{fig:diagnosis}d counts the steps
whose committed set has $\hat{R}(S) \ge \log \tau$ and $R(S) < \log \tau$.

Figure~\ref{fig:frontier}a and b average pass@1 and forward passes over
HumanEval, MBPP, GSM8K and MATH500, and c averages the two code benchmarks.
In panel a, PACT is faster than every published rule on every model whenever
the matched pass@1 lies between 50\% and 90\% of the best pass@1 of PACT.
Panel c compares PACT and fixed k at the same forward passes within the band of
11 to 26 forward passes per answer, and a light bar lies outside this band.
LLaDA2.1-mini is slower than the band at every setting, so c leaves it out. A
dark bar is significantly above one under a paired bootstrap over problems.

Figure~\ref{fig:analysis}a measures the dependence term $D(S)$ of each
committed set. We reveal the written tokens of $S$ one at a time in the order of
$\pi$, with one forward pass per token. The revealed probabilities give
$\log P_S(x_S)$, and $D(S)$ is this value minus $R(S)$. These forward passes
serve only the measurement and count toward no cost. Figure~\ref{fig:analysis}a
uses 40 HumanEval problems at $\tau = 0.05$, with 95\% intervals over problems.
It plots $|D(S)|$ divided by the size of $S$, over committed sets of at least
two positions.
Figure~\ref{fig:analysis}c decodes the code benchmarks at $\tau = 0.1$, and its
three rules run without spacing.

Figure~\ref{fig:hero} counts the answers in which a word of two or more letters
appears twice in a row. Each setting of a published decoder is paired with the
setting of PACT closest in forward passes on the same model and benchmark, when
the two lie within 15\%. The counts cover HumanEval, MBPP, GSM8K and MATH500 on
the four models. One token per step writes such a doubled word in about 1\% of
its answers.

\section{Further Results}
\label{app:further}

\subsection{Orders per Problem}
\label{app:orders}

\begin{figure}[t]
\centering
\includegraphics{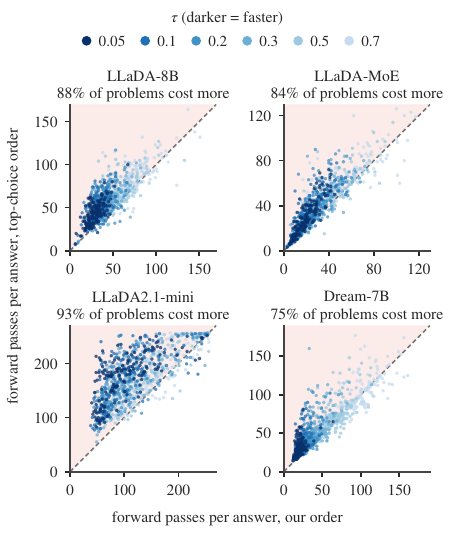}
\caption{\textbf{Forward passes per problem under two orders.} Each point is a
HumanEval problem at one $\tau$, under the top-choice order and under our order
with the same set-level cut. Each panel gives the share of problems in the
shaded triangle, where the top-choice order costs more.}
\label{fig:app_orders}
\end{figure}

Figure~\ref{fig:app_orders} compares the forward passes of each problem under
the top-choice order and under our order. Points above the diagonal are problems
on which the top-choice order costs more. On every model most problems lie above
it. This is the per-problem view of the cost in Figure~\ref{fig:diagnosis}e.

\subsection{32-Token Blocks}
\label{app:blocks}

Our experiments decode each answer as a single block of 256 positions, so the
commit rule alone decides which positions are filled first. Fast-dLLM decodes
in blocks of 32 tokens. Table~\ref{tab:blocks} repeats the comparison with
these blocks at temperature 0.6, on the code and math benchmarks of LLaDA-8B
and Dream-7B. PACT uses $g = 4$, since a block holds only
32 positions.

At the speed of Fast-dLLM, PACT reaches comparable accuracy on every benchmark.

\begin{table*}[t]
\centering
\small
\begin{tabular*}{\textwidth}{@{\extracolsep{\fill}}lrrrrrrrr}
\toprule
 & \multicolumn{4}{c}{LLaDA-8B} & \multicolumn{4}{c}{Dream-7B} \\
\cmidrule(lr){2-5}\cmidrule(lr){6-9}
 & HumanEval & MBPP & GSM8K & MATH500 & HumanEval & MBPP & GSM8K & MATH500 \\
\midrule
Speed of Fast-dLLM & 2.2$\times$ & 1.7$\times$ & 2.6$\times$ & 1.7$\times$ & 1.7$\times$ & 1.7$\times$ & 2.6$\times$ & 1.6$\times$ \\
Acc. of Fast-dLLM & 36.7 & 37.2 & 80.3 & 38.0 & 50.0 & 47.5 & 83.7 & 44.7 \\
Acc. of PACT & 39.3 & 38.1 & 80.9 & 37.5 & 51.5 & 47.4 & 85.3 & 44.6 \\
\bottomrule
\end{tabular*}
\caption{\textbf{Fast-dLLM and PACT with 32-token blocks.} Speed is tokens written per forward pass, and Acc. is pass@1 (\%) at the speed of Fast-dLLM. PACT is swept over $\tau$ with $g = 4$ and read at that speed by interpolation.}
\label{tab:blocks}
\end{table*}

\subsection{Decoding Time}
\label{app:wallclock}

\begin{figure*}[t]
\centering
\includegraphics{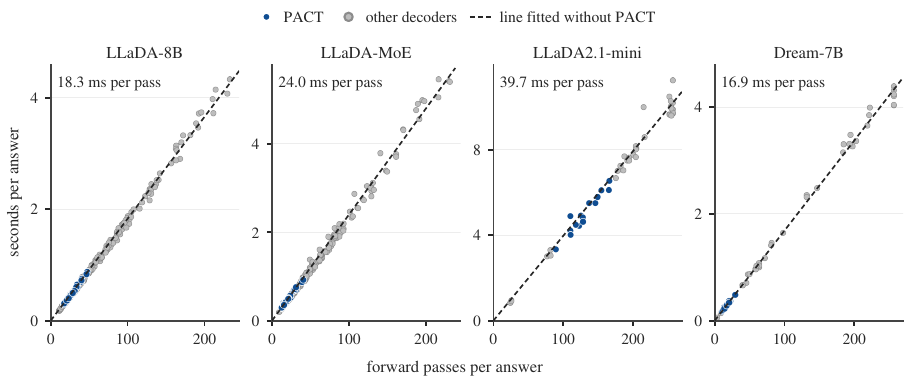}
\caption{\textbf{Decoding time against forward passes.} Each point is one
setting on the code benchmarks. The dashed line is fitted through the origin on
the settings of the other decoders, and its slope is printed as the time of one
forward pass.}
\label{fig:app_time}
\end{figure*}

Figure~\ref{fig:app_time} plots the seconds per answer of the settings in our
sweeps against their forward passes. On each model the points lie close to one
line through the origin, so the time of a forward pass hardly depends on the
commit rule. The points of PACT follow the line fitted without them, and the
time PACT saves follows the forward passes it saves.

\begin{figure}[t]
\centering
\includegraphics{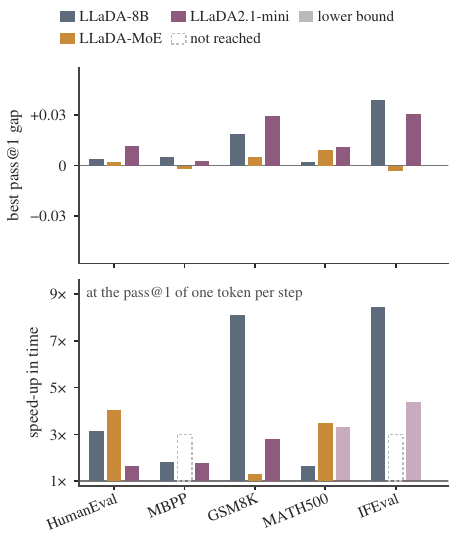}
\caption{\textbf{Speed-up in time at the accuracy of one token per step.} Top,
the best pass@1 of PACT minus that of one token per step. Bottom, the speed-up
at the fastest setting of PACT that reaches the pass@1 of one token per step. A
light bar is a lower bound, since the fastest setting already reaches it. A
dashed outline marks a benchmark where no setting reaches it.}
\label{fig:app_matched}
\end{figure}

Figure~\ref{fig:app_matched} matches PACT to the pass@1 of one token per step on
each benchmark of LLaDA-8B, LLaDA-MoE and LLaDA2.1-mini. PACT is faster on
almost every benchmark, by up to about eight times on LLaDA-8B.

\subsection{Autoregressive Model}
\label{app:ar}

Table~\ref{tab:ar} times decoding with one answer per problem. The diffusion
models decode greedily in 32-token blocks, as Fast-dLLM does, and every decoder
writes at most 256 tokens. PACT is faster than Qwen3-8B at both thresholds.

\begin{table}[t]
\centering
\small
\begin{tabular}{lrr}
\toprule
Decoder & HumanEval & GSM8K \\
\midrule
Qwen3-8B & 4.66 & 3.99 \\
\multicolumn{3}{l}{\quad LLaDA-8B} \\
\qquad One token per step & 5.50 (0.85$\times$) & 4.54 (0.88$\times$) \\
\qquad Fast-dLLM & 1.83 (2.6$\times$) & 1.34 (3.0$\times$) \\
\qquad PACT, $\tau = 0.3$ & \underline{1.02 (4.6$\times$)} & \underline{0.72 (5.5$\times$)} \\
\qquad PACT, $\tau = 0.05$ & \textbf{0.65 (7.2$\times$)} & \textbf{0.49 (8.1$\times$)} \\
\multicolumn{3}{l}{\quad Dream-7B} \\
\qquad One token per step & 5.24 (0.89$\times$) & 4.57 (0.87$\times$) \\
\qquad Fast-dLLM & 2.15 (2.2$\times$) & 1.36 (2.9$\times$) \\
\qquad PACT, $\tau = 0.3$ & \underline{1.31 (3.6$\times$)} & \underline{0.78 (5.1$\times$)} \\
\qquad PACT, $\tau = 0.05$ & \textbf{0.64 (7.3$\times$)} & \textbf{0.51 (7.8$\times$)} \\
\bottomrule
\end{tabular}
\caption{\textbf{Time per problem against an autoregressive model.} Seconds per problem with one answer per problem, and in parentheses the speed-up over Qwen3-8B. All decoders write at most 256 tokens on the same GPU, on 50 problems of each benchmark. The diffusion models decode greedily in 32-token blocks, and Qwen3-8B decodes with a key-value cache. Fastest within a model in bold, second fastest underlined.}
\label{tab:ar}
\end{table}

\subsection{Temperature on GSM8K}
\label{app:temperature}

\begin{figure}[t]
\centering
\includegraphics{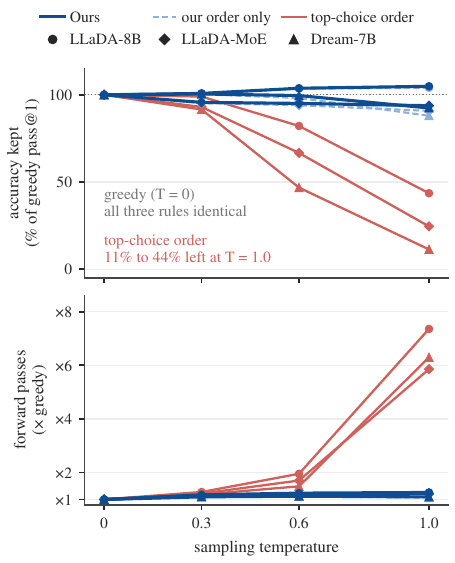}
\caption{\textbf{Sampling temperature on GSM8K.} Pass@1 as a share of greedy
pass@1 (top) and forward passes as a multiple of greedy (bottom), for the three
rules of Figure~\ref{fig:analysis}c on three models.}
\label{fig:app_temperature}
\end{figure}

Figure~\ref{fig:app_temperature} repeats Figure~\ref{fig:analysis}c on GSM8K.
As the temperature rises, the top-choice order loses accuracy and spends more
forward passes. PACT changes far less on both counts.

\subsection{Notes on Section 4}
\label{app:notes}

Over the eight answers of a problem, PACT gives about as many distinct final
answers as one token per step on GSM8K and MATH500, and a similar share of
distinct 4-grams on every benchmark.

Without the written-word score, committed sets that report more than they write
grow more common as $\tau$ falls (Figure~\ref{fig:ablation}b). Faster steps
commit more positions, and less certain ones.

The saving of spacing holds on every task group of
Figure~\ref{fig:ablation}c, including code answers of 512 tokens, twice the
usual length. On IFEval, spacing also raises pass@1 at every threshold we ran.

The three rules of Figure~\ref{fig:analysis}c write the same tokens under greedy
decoding. They diverge once the drawn token can differ from the most likely one,
which is where ordering by the top choice starts to lose accuracy.

\end{document}